\pdfoutput=1

\documentclass[11pt]{article}
\usepackage{acl2015}
\usepackage{times}
\usepackage{url}
\usepackage{latexsym}

\usepackage{multirow}
\usepackage{graphicx}
\usepackage{subfig}
\usepackage{amsmath}
\usepackage{verbatim}

\title{Contrastive Analysis with Predictive Power: Typology Driven Estimation of Grammatical Error Distributions in ESL}

\author{Yevgeni Berzak \\
  CSAIL MIT \\
  {\tt berzak@mit.edu} \\\And
  Roi Reichart \\
  Technion IIT \\
  {\tt roiri@ie.technion.ac.il} \\\And
  Boris Katz \\
  CSAIL MIT \\
  {\tt boris@mit.edu} \\}

\date{}

\begin{document}
\maketitle
\begin{abstract}

This work examines the impact of cross-linguistic transfer on 
grammatical errors in English as Second Language (ESL) texts. 
Using a computational framework that formalizes the theory of Contrastive
Analysis (CA), we demonstrate that \emph{language specific 
error distributions} in ESL writing can be predicted from the typological 
properties of the native language and their relation to the typology
of English. Our typology driven model enables to obtain accurate estimates of 
such distributions without access to any ESL data for the target languages. 
Furthermore, we present a strategy for adjusting our method
to low-resource languages that lack typological documentation using a bootstrapping 
approach which approximates native language typology from ESL texts. Finally, we show that our 
framework is instrumental for linguistic inquiry seeking to identify first language factors 
that contribute to a wide range of difficulties in second language acquisition. 

\end{abstract}

\section{Introduction}

The study of cross-linguistic transfer, whereby properties of a native 
language influence performance in a foreign language, has a long tradition 
in Linguistics and Second Language Acquisition (SLA). Much of the linguistic
work on this topic was carried out within the framework of Contrastive Analysis (CA), 
a theoretical approach that aims to explain difficulties in second language 
learning in terms of the relations between structures in the native and 
foreign languages. 

The basic hypothesis of CA was formulated by Lado \shortcite{lado1957}, 
who suggested that ``we can predict and describe the patterns that will 
cause difficulty in learning, and those that will not cause difficulty, 
by comparing systematically the language and culture to be learned with 
the native language and culture of the student''. In particular, Lado 
postulated that divergences between the native and foreign languages will 
negatively affect learning and lead to increased error rates in the foreign 
language. This and subsequent hypotheses were soon met with criticism, 
targeting their lack of ability to provide reliable predictions, leading to 
an ongoing debate on the extent to which foreign language 
errors can be explained and predicted by examining native language structure.

Differently from the SLA tradition, which emphasizes manual analysis of 
error case studies \cite{odlin1989transfer}, we address the heart of this 
controversy from a computational data-driven perspective, focusing on the 
issue of predictive power. We provide a formalization of the CA framework, 
and demonstrate that the relative frequency of grammatical 
errors in ESL can be reliably predicted from the typological properties 
of the native language and their relation to the typology of English 
using a regression model. 

Tested on 14 languages in a leave-one-out fashion, our model achieves
a Mean Average Error (MAE) reduction of 21.8\% in predicting the language specific relative 
frequency of the 20 most common ESL structural error types, as compared to 
the relative frequency of each of the error types in the training data,
yielding improvements across all the languages and the large majority of the error types.
Our regression model also outperforms a stronger, nearest neighbor based baseline, that projects
the error distribution of a target language from its typologically closest language.

While our method presupposes the existence of typological annotations for the 
test languages, we also demonstrate its viability in low-resource scenarios for 
which such annotations are not available. To address this setup, we present a bootstrapping 
framework in which the typological features required for prediction of grammatical errors
are approximated from automatically extracted ESL morpho-syntactic features using the method of \cite{berzak2014}. 
Despite the noise introduced in this process, our bootstrapping strategy achieves an
error reduction of 13.9\% compared to the average frequency baseline. 

Finally, the utilization of typological features as predictors, 
enables to shed light on linguistic factors that could give rise 
to different error types in ESL. For example, in accordance with common linguistic
knowledge, feature analysis of the model suggests that the main contributor 
to increased rates of determiner omission in ESL is the lack of determiners in 
the native language. A more complex case of missing pronouns is intriguingly tied
by the model to native language subject pronoun marking on verbs.

To summarize, the main contribution of this work is a CA inspired computational
framework for learning language specific grammatical error distributions in ESL. Our approach is 
both predictive and explanatory. It enables us to obtain improved estimates for 
language specific error distributions without access to ESL error annotations for the target 
language. Coupling grammatical errors with typological information also provides 
meaningful explanations to some of the linguistic factors that drive the 
observed error rates. 

The paper is structured as follows. Section \ref{sec:related-work} surveys
related linguistic and computational work on cross-linguistic transfer. Section 
\ref{sec:data} describes the ESL corpus and the typological data used in 
this study. In section \ref{sec:var-analysis} we motivate our native language
oriented approach by providing a variance analysis for ESL errors across native 
languages. Section \ref{sec:prediction} presents the regression model for 
prediction of ESL error distributions. The bootstrapping framework which utilizes 
automatically inferred typological features is described in section 
\ref{sec:bootstrapping}. Finally, we present the conclusion and directions 
for future work in section \ref{sec:conclusion}.

\section{Related Work}\label{sec:related-work}

Cross linguistic-transfer was extensively studied in SLA, Linguistics and Psychology 
\cite{odlin1989transfer,gass1992language,jarvis2007transfer}. Within this area
of research, our work is most closely related to the Contrastive Analysis (CA) framework. 
Rooted in the comparative linguistics tradition, CA was first suggested by Fries 
\shortcite{fries1945} and formalized by Lado \shortcite{lado1957}. In essence, CA 
examines foreign language performance, with a particular focus on learner difficulties, 
in light of a structural comparison between the native and the foreign languages. 
From its inception, CA was criticized for the lack of a solid predictive theory
\cite{wardhaugh1970,whitman1972unpredictability}, leading to an ongoing scientific debate 
on the relevance of comparison based approaches. Important to our study is that the type of evidence used in this 
debate typically relies on small scale manual case study analysis. Our work seeks to 
reexamine the issue of predictive power of CA based methods using a computational, 
data-driven approach.

Computational work touching on cross-linguistic transfer was mainly conducted in 
relation to the Native Language Identification (NLI) task, in which the 
goal is to determine the native language of the author of an ESL text. Much of this work 
focuses on experimentation with different feature sets 
\cite{nli2013report}, including features derived from the CA framework \cite{wong2009contrastive}. 
A related line of inquiry which is closer to our work deals with
the identification of ESL syntactic patterns that are specific to speakers of different native languages \cite{swanson2013,swanson2014}. 
Our approach differs from this research direction by focusing on grammatical errors,
and emphasizing prediction of language specific patterns rather than their identification.

Previous work on grammatical error correction that examined determiner and preposition errors \cite{rozovskaya2011,rozovskaya2014} 
incorporated native language specific priors in models that are otherwise trained on standard 
English text. Our work extends the native language tailored treatment of grammatical errors to 
a much larger set of error types. More importantly, this approach is limited by the 
availability of manual error annotations for the target language in order to obtain the 
required error counts. Our framework enables to bypass this annotation bottleneck by 
predicting language specific priors from typological information.

The current investigation is most closely related to studies that
demonstrate that ESL signal can be used to infer pairwise similarities
between native languages \cite{nagata2013,berzak2014} and in particular,
tie the similarities to the typological characteristics of these languages \cite{berzak2014}.
Our work inverts the direction of this analysis by 
starting with typological features, and utilizing them to predict 
error patterns in ESL. We also show that the two approaches can be 
combined in a bootstrapping strategy by first inferring typological properties 
from automatically extracted morpho-syntactic ESL features, and in turn, 
using these properties for prediction of language specific error distributions in ESL.

\section{Data}\label{sec:data}

\subsection{ESL Corpus}

We obtain ESL essays from the Cambridge First Certificate in English (FCE) 
learner corpus \cite{fcecorpus2011}, a publicly available subset of the Cambridge Learner Corpus 
(CLC)\footnote{\url{http://www.cambridge.org/gb/elt/catalogue/subject/custom/item3646603}}.
The corpus contains upper-intermediate level
essays by native speakers of 16 languages\footnote{We plan to extend our analysis to 
additional proficiency levels and languages when error annotated data for these learner profiles will be publicly available.}. 
Discarding Swedish and Dutch, which 
have only 16 documents combined, we take into consideration the remaining following 14 languages, 
with the corresponding number of documents in parenthesis: Catalan (64), Chinese (66), 
French (146), German (69), Greek (74), Italian (76), Japanese (82), Korean (86), 
Polish (76), Portuguese (68), Russian (83), Spanish (200), Thai (63) and Turkish (75). 
The resulting dataset contains 1228 documents with an average of 379 words per document. 

The FCE corpus has an elaborate error annotation scheme \cite{nicholls2003clc} 
and high quality of error annotations, making it particularly suitable for
our investigation. The annotation scheme encompasses 75 
different error types, covering a wide range of grammatical errors on different 
levels of granularity. 
As the typological features used in this work refer mainly to \emph{structural}
properties, we filter out spelling errors, punctuation errors and open class semantic errors,
remaining with a list of grammatical errors that are typically related to language structure. 
We focus on the 20 most frequent error types\footnote{Filtered errors that would have otherwise appeared
in the top 20 list, with their respective rank in brackets: 
Spelling (1), Replace Punctuation (2), Replace Verb (3), Missing Punctuation (7), Replace (8), Replace Noun (9) 
Unnecessary Punctuation (13), Replace Adjective (18), Replace Adverb (20).} in this list, which are presented and exemplified in table \ref{FCE_errors}.
In addition to concentrating on the most important structural ESL errors, this 
cutoff prevents us from being affected by data sparsity issues associated 
with less frequent errors.

\begin{table*}[!ht]
\footnotesize
\begin{center}
\begin{tabular}{|l|l|l|l|l|l|l|}
\hline
\bf Rank &\bf Code & \bf Name            & \bf Example & \bf Count  & \bf KW & \bf MW \\ \hline
1&TV       & Verb Tense          & I hope I \textbf{give} \emph{have given} you enough details & 3324 & ** & 34 \\ \hline
2&RT       & Replace Preposition & \textbf{on} \emph{in} July                                  & 3311 & ** & 31 \\ \hline
3&MD       & Missing Determiner  & I went for \emph{the} interview                             & 2967 & ** & 57 \\ \hline
4&FV       & Wrong Verb Form     & had time to \textbf{played} \emph{play}                     & 1789 & ** & 21 \\ \hline
5&W        & Word Order          &  \textbf{Probably} our homes will \emph{probably} be & 1534 & ** & 34 \\ \hline
6&MT       & Missing Preposition &  explain \emph{to} you                                      & 1435 & ** & 22 \\ \hline
7&UD       & Unnecessary Determiner  & a course at \textbf{the} Cornell University             & 1321 &    &       \\ \hline
8&UT       & Unnecessary Preposition & we need it \textbf{on} each minute                      & 1079 &    &       \\ \hline
9&MA       & Missing Pronoun     & because \emph{it} is the best conference                   & 984  & ** & 33 \\ \hline
10&AGV      & Verb Agreement      & the teachers \textbf{was} \emph{were} very experienced      & 916  & ** & 21 \\ \hline
11&FN       & Wrong Form Noun     & because of my \textbf{study} \emph{studies}                 & 884  & ** & 24 \\ \hline
12&RA       & Replace Pronoun     & she just met Sally, \textbf{which} \emph{who}               & 847  & ** & 17 \\ \hline
13&AGN      & Noun Agreement      & two \textbf{month} \emph{months} ago                        & 816  & ** & 24 \\ \hline
14&RD       & Replace Determiner  &  of \textbf{a} \emph{the} last few years                    & 676  & ** & 35\\ \hline
15&DJ       & Wrongly Derived Adjective & The mother was \textbf{pride} \emph{proud}            & 608  & *  & 8 \\ \hline
16&DN       & Wrongly Derived Noun      &  \textbf{working place} \emph{workplace}              & 536  &   & \\ \hline
17&DY       & Wrongly Derived Adverb    &  \textbf{Especial} \emph{Especially}                  & 414  & ** & 14 \\ \hline
18&UA       & Unnecessary Pronoun       & feel \textbf{ourselves} comfortable                   & 391  & *  & 9 \\ \hline
19&MC       & Missing Conjunction       &  reading, \emph{and} playing piano at home            &  346 & *  & 11 \\ \hline
20&RC       & Replace Conjunction       &  not just the car, \textbf{and} \emph{but} also the train & 226 &  &  \\ \hline

\end{tabular}
\end{center}
\caption{The 20 most frequent error types in the FCE corpus that are related to language structure. 
In the Example column, words marked in italics are corrections for the words marked in bold. 
The Count column lists the overall count of each error type in the corpus.
The KW column depicts the result of the Kruskal-Wallis test whose null hypothesis is that the relative error frequencies 
for different native languages are drawn from the same distribution. 
Error types for which this hypothesis is rejected with $p < 0.01$ are denoted with `*'. Error types with
$p < 0.001$ are marked with `**'. The MW column denotes the number of language pairs (out of the total 91 pairs)
which pass the post-hoc Mann-Whitney test with $p < 0.01$.}\label{FCE_errors}
\end{table*}

\subsection{Typological Database}\label{subsec:typo}

We use the World Atlas of Language Structures (WALS; Dryer and Haspelmath, 2013), a repository of typological 
features of the world's languages, as our source of 
linguistic knowledge about the native languages of the ESL corpus authors.
The features in WALS are divided into 11 categories: 
Phonology, Morphology, Nominal Categories, Nominal Syntax, Verbal Categories, Word Order, 
Simple Clauses, Complex Sentences, Lexicon, Sign Languages and Other. 
Table \ref{wals-examples} presents examples of WALS features belonging to different categories. 
The features can be associated with different variable types, 
including binary, categorical and ordinal, making their encoding a challenging task. 
Our strategy for addressing this issue is feature binarization (see section \ref{subsec:prediction-features}).

An important challenge introduced by the WALS database is incomplete documentation. 
Previous studies \cite{daume2009clustering,georgi2010clustering} have estimated that 
only 14\% of all the language-feature combinations in the database have 
documented values. While this issue is most acute for low-resource languages, even 
the well studied languages in our ESL dataset are lacking a significant portion of 
the feature values, inevitably hindering the effectiveness of our approach. 

We perform several preprocessing steps in order to select the features that will
be used in this study. First, as our focus is on structural features that can be expressed in written form, we discard all the
features associated with the categories Phonology, Lexicon\footnote{The discarded Lexicon features
refer to properties such as the number of words in the language that denote colors, and identity of word pairs such as ``hand'' and ``arm''.}, Sign Languages and Other.
We further discard 24 features which either have a documented value for only one language,
or have the same value in all the languages.  
The resulting feature-set contains 119 features, with an average of 2.9
values per feature, and 92.6 documented features per language.

{\renewcommand{\arraystretch}{1.1}
\begin{table}[!ht]
\small
\begin{center}
\begin{tabular}{|l|l|l|l|}
\hline
\bf ID & \bf Category           & \bf  Name            & \bf Values  \\ \hline
23A    & Morphology         & Locus of             & No case marking,   \\
       &                    & Marking                 & Core cases only, \\ 
       &                    & in the 	        & Core and non-core, \\ 
	&	            &	Clause		&No syncretism \\ \hline
67A    & Verbal             & The Future                & Inflectional future,  \\ 
       & Categories         &     Tense                  & No inflectional  \\ 
	&			&			& future. \\ \hline
30A    & Nominal            & Number of                   & None, Two, Three, \\
       & Categories         & Genders                     & Four, Five or more. \\ \hline
87A    & Word Order         & Order of                    & AN, NA, No  \\
	&	            & Adjective               & dominant order, \\
	&	            & and Noun			&Only internally \\
       &                    &                           &headed relative \\ 
	&			&			&clauses. \\ \hline
\end{tabular}
\end{center}
\caption{Examples of WALS features. 
	}\label{wals-examples}
\end{table}
}

\section{Variance Analysis of Grammatical Errors in ESL}\label{sec:var-analysis}

To motivate a native language based treatment of grammatical error distributions
in ESL, we begin by examining whether there is a statistically significant 
difference in ESL error rates based on the native language of the learners. 
This analysis provides empirical justification for our approach, 
and to the best of our knowledge was not conducted in previous studies.

To this end, we perform a Kruskal-Wallis (KW) test \cite{kruskal1952} for 
each error type\footnote{We chose the non-parametric KW rank-based test over ANOVA, 
as according to the Shapiro-Wilk \shortcite{shapiro1965} and Levene \shortcite{levene1960} 
tests, the assumptions of normality and homogeneity of variance do not hold 
for our data. In practice, the ANOVA test yields similar results to those 
of the KW test.}. We treat the relative error frequency per word
in each document as a sample\footnote{We also performed the KW test on the absolute 
error frequencies (i.e. raw counts) per word, obtaining similar results to the ones reported here on the 
relative frequencies per word.} (i.e. the relative frequencies of all the error types in a document sum to 1). 
The samples are associated with 14 groups, according to the native language of the 
document's author. For each error type, the null hypothesis of the test is that error fraction samples 
of all the native languages are drawn from the same underlying distribution. 
In other words, rejection of the null hypothesis implies a significant difference 
between the relative error frequencies of at least one language pair.

As shown in table \ref{FCE_errors}, we can reject the null hypothesis for 
16 of the 20 grammatical error types with $p < 0.01$, where
Unnecessary Determiner, Unnecessary Preposition, Wrongly Derived Noun, and 
Replace Conjunction are the error types that do not exhibit dependence on 
the native language. 
Furthermore, the null hypothesis can be rejected for 13 error types with $p < 0.001$. 
These results suggest that the relative error rates of the majority of the common 
structural grammatical errors in our corpus indeed differ between native speakers 
of different languages. 

We further extend our analysis by performing pairwise post-hoc Mann-Whitney (MW) 
tests \cite{mann1947test} in order to determine the number of language pairs that
significantly differ with respect to their native speakers' error fractions in ESL. Table 
\ref{FCE_errors} presents the number of language pairs that pass this test with 
$p < 0.01$ for each error type. This inspection suggests Missing Determiner as 
the error type with the strongest dependence on the author's native language, followed
by Replace Determiner, Verb Tense, Word Order, Missing Pronoun and Replace Preposition. 

\section{Predicting Language Specific Error Distributions in ESL}\label{sec:prediction}

\subsection{Task Definition}
Given a language $l \in L$, our task is to predict for this language the relative error 
frequency $y_{l,e}$ of each error type $e \in E$, where $L$ is the set of all native languages, 
$E$ is the set of grammatical errors, and $\sum_{e} y_{l,e} = 1$.

\subsection{Model}\label{subsec:prediction-model}

In order to predict the error distribution of a native language, we train regression models
on individual error types: 
\begin{equation}
\hat{y}'_{l,e} = \theta_{l,e} \cdot f(t_{l}, t_{eng})
\end{equation}
In this equation $\hat{y}'_{l,e}$ is the predicted relative frequency of an error
of type $e$ for ESL documents authored by native speakers of language $l$, 
and $f(t_{l}, t_{eng})$ is a feature vector derived from the typological 
features of the native language $t_{l}$ and the typological features of English $t_{eng}$.

The model parameters $\theta_{l,e}$ are obtained 
using Ordinary Least Squares (OLS) on the training data $D$, 
which consists of typological feature vectors paired with relative error 
frequencies of the remaining 13 languages:
\begin{equation}
D = \{(f(t_{l'}, t_{eng}), y_{e,l'}) | l' \in L, l' \neq l \}
\end{equation}
To guarantee that the individual relative error frequency estimates sum to 1 for each language, we
renormalize them to obtain the final predictions: 
\begin{equation}
\hat{y}_{l,e} = \frac{\hat{y}'_{l,e}}{\sum_{e} \hat{y}'_{l,e}}
\end{equation}

\subsection{Features}\label{subsec:prediction-features}
Our feature set can be divided into two subsets.
The first subset, used in a version of our model called \emph{Reg}, contains the typological features of the native language.
In a second version of our model, called \emph{RegCA}, we also utilize additional features that 
explicitly encode differences between the typological features of the native language, and the 
and the typological features of English.

\subsubsection{Typological Features}
In the \emph{Reg} model, we use the typological features of the native language that are documented in WALS.
As mentioned in section \ref{subsec:typo}, WALS features belong to different variable
types, and are hence challenging to encode. We address this issue by binarizing all the features. 
Given $k$ possible values $v_{k}$ for a given WALS feature $t_{i}$, we generate 
$k$ binary typological features of the form:
\begin{multline}
f_{i,k}(t_{l}, t_{eng}) = \begin{cases}
	   			1 & \text{ if } t_{l, i} = v_{k} \\
                        	0 & \text{ otherwise}
                          \end{cases}
\end{multline}
When a WALS feature of a given language does not have a documented value, all $k$ entries
of the feature for that language are assigned the value of 0. This process transforms the original 119 WALS features
into 340 binary features. 

\subsubsection{Divergences from English}
In the spirit of CA, in the model \emph{RegCA}, we also utilize features that explicitly encode differences between 
the typological features of the native language and those of English. 
These features are also binary, and take the value 1 when the value of a WALS feature 
in the native language is different from the corresponding value in English: 
\begin{multline}
f_{i}(t_{l}, t_{eng}) = \begin{cases}
	   			1 & \text{ if } t_{l,i} \neq t_{eng,i}  \\
                        	0 & \text{ otherwise}
                          \end{cases}
\end{multline}
We encode 104 such features, in accordance with the typological features of English 
available in WALS. The features are activated only when a typological
feature of English has a corresponding documented feature in the native language. The
addition of these divergence features brings the total number of features in our 
feature set to 444. 

\subsection{Results}\label{sec:results-reg}

We evaluate the model predictions using two metrics. The first
metric, Absolute Error, measures the distance between the predicted and the true relative frequency of 
each grammatical error type\footnote{For clarity of presentation, all the reported results on this metric are multiplied by 100.}:
\begin{equation}
\text{Absolute Error} = |\hat{y}_{l,e} - y_{l,e}|
\end{equation}  
When averaged across different predictions we refer to this metric as Mean Absolute Error (MAE).
 
The second evaluation score is the Kullback-Leibler divergence $D_{KL}$, 
a standard measure for evaluating the difference between two distributions. 
This metric is used to evaluate the predicted grammatical error distribution of a native language:
\begin{equation}
D_{KL} (y_{l}||\hat{y}_{l}) = \sum_{e}{ y_{l,e} \ln \frac{y_{l,e}}{\hat{y}_{l,e}}}
\end{equation}   

{\renewcommand{\arraystretch}{1.1}
\begin{table}[ht!]
\footnotesize
\begin{center}
\begin{tabular}{|l|l|l|l|l|}
\hline
\bf                & \bf Base & \bf NN & \bf Reg & \bf RegCA  \\ \hline
 \textbf{MAE}  &1.28 & 1.11 & 1.02 & \textbf{1.0}  \\ \hline 
 \textbf{Error Reduction} &   -  & 13.3 & 20.4  & \textbf{21.8} \\ \hline 
 \#Languages   &   -      & 9/14 & 12/14 & \textbf{14/14} \\ \hline 
 \#Mistakes    &   -      & 11/20& \textbf{15/20} &  14/20 \\ \hline \hline
\textbf{AVG $\mathbf{D_{KL}}$}  & 0.052 & 0.046 & 0.033 & \textbf{0.032} \\ \hline 
 \#Languages   &   -  & 10/14 & \textbf{14/14}& \textbf{14/14} \\ \hline 
  
\end{tabular}
\end{center}
\caption{Results for prediction of relative error frequencies using the \emph{MAE} metric across languages and error types,
 and the $D_{KL}$ metric averaged across languages.
	\emph{\#Languages} and \emph{\#Mistakes} denote the number of languages and grammatical error types on which a model outperforms 
	\emph{Base}. 
	}\label{results_summary}
\end{table}
}

Table \ref{results_summary} summarizes the grammatical error prediction results\footnote{As described 
in section \ref{subsec:prediction-model}, we report the performance 
of regression models trained and evaluated on relative error frequencies 
obtained by normalizing the rates of the different error types.
We also experimented with training and evaluating the models on absolute error counts  
per word, obtaining results that are similar to those reported here.}.
The baseline model \emph{Base} sets the relative frequencies of the grammatical 
errors of a test language to the respective relative error frequencies in the training 
data. We also consider a stronger, language specific model called \emph{Nearest Neighbor (NN)}, 
which projects the error distribution of a target language from the typologically closest language 
in the training set, according to the cosine similarity measure. This baseline 
provides a performance improvement for the majority of the languages and error
types, with an average error reduction of 13.3\% on the \emph{MAE}
metric compared to \emph{Base}, and improving from 0.052 to 0.046 on the 
KL divergence metric, thus emphasizing the general advantage 
of a native language adapted approach to ESL error prediction. 

Our regression model introduces further substantial performance improvements. 
The \emph{Reg} model, which uses the typological features of the native 
language for predicting ESL relative error frequencies, achieves 20.4\% 
\emph{MAE} reduction over the \emph{Base} model. The \emph{RegCA} version of the 
regression model, which also incorporates differences between the 
typological features of the native language and English, surpasses the \emph{Reg} model, 
reaching an average error reduction of 21.8\% from the \emph{Base} model, 
with improvements across all the languages and the majority of the error types.
Strong performance improvements are also obtained on the KL divergence measure, 
where the \emph{RegCA} model scores 0.032, compared to the baseline score of 0.052.

To illustrate the outcome of our approach, consider the example in table 
\ref{jap_example}, which compares the top 10 predicted errors for Japanese using the  
\emph{Base} and \emph{RegCA} models. In this example, \emph{RegCA} correctly places Missing
Determiner as the most common error in Japanese, with a significantly higher relative frequency 
than in the training data. Similarly, it provides an accurate prediction for the Missing 
Preposition error, whose frequency and rank are underestimated by the \emph{Base} model. 
Furthermore, \emph{RegCA} correctly predicts the frequency of Replace Preposition and Word 
Order to be lower than the average in the training data.

\begin{table*}[ht]
\footnotesize
\begin{center}
\begin{tabular}{|l|l|l|l|l|l|l|}
\hline
\bf Rank & \bf Base & \bf Frac. & \bf RegCA & \bf Frac. &\bf True  & \bf Frac.  \\ \hline
\bf 1 & \textbf{Replace Preposition} & \textbf{0.14}         & \textbf{Missing Determiner} & \textbf{0.18} & \textbf{Missing Determiner} & \textbf{0.20} \\ \hline
\bf 2 & Tense Verb          & 0.14         &  Tense Verb         & 0.12                  & Tense Verb & 0.12 \\ \hline
\bf 3 & \textbf{Missing Determiner} & \textbf{0.12} &  \textbf{Replace Preposition} & \textbf{0.12}                 &  \textbf{Replace Preposition} &  \textbf{0.10} \\ \hline
\bf 4 & Wrong Verb Form & 0.07    & \textbf{Missing Preposition} & \textbf{0.08}& \textbf{Missing Preposition} & \textbf{0.08} \\ \hline
\bf 5 & \textbf{Word Order} & \textbf{0.06}                  & Unnecessary Determiner & 0.06 & Unnecessary Preposition & 0.06 \\ \hline 
\bf 6 & \textbf{Missing Preposition} & \textbf{0.06}         & Wrong Verb Form  & 0.05             & Unnecessary Determiner & 0.05 \\ \hline 
\bf 7 & Unnecessary Determiner & 0.06      & Unnecessary Preposition & 0.05     & Replace Determiner & 0.05 \\ \hline 
\bf 8 & Unnecessary Preposition & 0.04     & Wrong Noun Form & 0.05             & Wrong Verb Form & 0.05 \\ \hline 
\bf 9 & Missing Pronoun & 0.04             & \textbf{Word Order} & \textbf{0.05} & \textbf{Word Order} & \textbf{0.04} \\ \hline 
\bf 10 &Wrong Noun Form & 0.04             & Verb Agreement & 0.04 	 & Wrong Noun Form & 0.06 \\ \hline 
\end{tabular}
\end{center}
\caption{Comparison between the fractions and ranks of the top 10 predicted error types by the \emph{Base} and 
	\emph{RegCA} models for Japanese. As opposed to the \emph{Base} method, the 
	\emph{RegCA} model correctly predicts Missing Determiner to be the most frequent 
	error committed by native speakers of Japanese. It also correctly predicts Missing Preposition
	to be more frequent and Replace Preposition and Word Order to be less frequent than
	in the training data.}\label{jap_example}
\end{table*}

\subsection{Feature Analysis}\label{subsec:results-feat-analysis}

An important advantage of our typology-based approach are the clear semantics
of the features, which facilitate the interpretation of the model.
Inspection of the model parameters allows us to gain insight into the typological 
features that are potentially involved in causing different types of ESL errors. 
Although such inspection is unlikely to provide a comprehensive coverage of all the relevant
causes for the observed learner difficulties, it can serve as a valuable starting point for exploratory linguistic analysis and
formulation of a cross-linguistic transfer theory. 

Table \ref{features} lists the most salient typological features, as determined 
by the feature weights averaged across the models of different languages, for the error 
types Missing Determiner and Missing Pronoun. In the case of determiners, the model  
identifies the lack of definite and indefinite articles in the native language 
as the strongest factors related to increased rates of determiner omission. Conversely, 
features that imply the presence of an article system in the native language, such 
as `Indefinite word same as ’one’' and `Definite word distinct from demonstrative' 
are indicative of reduced error rates of this type.

A particularly intriguing example concerns the Missing Pronoun error. The most 
predictive typological factor for increased pronoun omissions is
pronominal subject marking on the verb in the native language. Differently from 
the case of determiners, it is not the lack of the relevant structure in the native 
language, but rather its different encoding that seems to drive erroneous pronoun omission. 
Decreased error rates of this type correlate most strongly with
obligatory pronouns in subject position, as well as a verbal person marking system similar 
to the one in English.

{\renewcommand{\arraystretch}{1.1}
\begin{table}[ht!]
\scriptsize
\begin{center}
\begin{tabular}{|l|l|}

\hline
        \multicolumn{2}{|c|}{\textbf{Missing Determiner}}   \\ \hline
        \textbf{37A Definite Articles: Different from English} 			& .057    \\ \hline
	\textbf{38A Indefinite Articles: No definite or indefinite article}     & .055 \\ \hline
	\textbf{37A Definite Articles: No definite or indefinite article}   & .055 \\ \hline 
		49A Number of Cases: 6-7 case  & .052 \\ \hline \hline 

	100A Alignment of Verbal Person Marking: Accusative & -.073    \\ \hline
        \textbf{38A Indefinite Article: Indefinite word same as 'one'}     & -.050    \\ \hline
        52A Comitatives and Instrumentals: Identity      & -.044    \\ \hline
        \textbf{37A Definite Articles:}                           & -.036    \\ 
        \textbf{Definite word distinct from demonstrative} & \\ \hline \hline

        \multicolumn{2}{|c|}{\textbf{Missing Pronoun}} \\ \hline
        \textbf{101A Expression of Pronominal Subjects:}   	& .015    \\
	\textbf{Subject affixes on verb}                         & \\ \hline 
   	71A The Prohibitive:   Different from English  & .012    \\ \hline
	38A Indefinite Articles: Indefinite word same as 'one'      & .011    \\ \hline 
	71A The Prohibitive: Special imperative + normal negative  & .010    \\ \hline \hline
	\textbf{104A Order of Person Markers on the Verb:}      & -.016    \\ 
        \textbf{A \& P do not or do not both occur on the verb}& \\ \hline
        \textbf{102A Verbal Person Marking: Only the A argument} & -.013    \\  \hline
	\textbf{101A Expression of Pronominal Subjects:}  & -.011    \\ 
	\textbf{Obligatory pronouns in subject position} & \\ \hline
        71A The Prohibitive: Normal imperative + normal negative       & -.010    \\ \hline
        
\end{tabular}
\end{center}
\caption{The most predictive typological features of the \emph{RegCA} model for the errors
Missing Determiner and Missing Pronoun. The right column depicts the feature weight 
averaged across all the languages. Missing determiners are related to the absence of a determiner 
system in the native language. Missing pronouns are correlated with subject pronoun marking on the verb.}\label{features}
\end{table}
}

\section{Bootstrapping with ESL-based Typology}\label{sec:bootstrapping}

Thus far, we presupposed the availability of substantial typological information 
for our target languages in order to predict their ESL error distributions. 
However, the existing typological documentation for the 
majority of the world's languages is scarce, limiting the applicability of this 
approach for low-resource languages.

We address this challenge for scenarios in which an unannotated collection of 
ESL texts authored by native speakers of the target language is available. 
Given such data, we propose a bootstrapping strategy which uses the method 
proposed in \cite{berzak2014} in order to approximate the typology of the native 
language from morpho-syntactic features in ESL.
The inferred typological features serve, in turn, as a proxy for the true 
typology of that language in order to predict its speakers' ESL grammatical error 
rates with our regression model.

To put this framework into effect, we use the FCE corpus to train a log-linear model for native language
classification using morpho-syntactic features obtained from the output of
the Stanford Parser \cite{2006stanforddep}:
\begin{equation}
p(l|x;\theta) = \frac{\exp(\theta \cdot f(x,l))}{\sum_{l' \in L} \exp(\theta \cdot f(x,l'))}
\end{equation}
where $l$ is the native language, $x$ is the observed English document and 
$\theta$ are the model parameters. 
We then derive pairwise similarities between languages by
averaging the uncertainty of the model with respect to each language pair:
\begin{multline}
S'_{ESL_{l,l'}} = 
	\begin{cases}
	\frac{1}{\left\vert{D_{l}}\right\vert}\sum\limits_{(x, l) \in D_{l}} p(l'|x;\theta) & \text{if } l' \neq l \\
        1 & \text{\small otherwise}
        \end{cases}
\end{multline}
In this equation, $x$ is an ESL document, $\theta$ are the parameters of the native language classification model and 
$D_{l}$ is a set of documents whose native language is $l$. For each pair of languages $l$
and $l'$ the matrix $S'_{ESL}$ contains an entry $S'_{ESL_{l,l'}}$ which represents 
the average probability of confusing $l$ for $l'$, and an entry $S'_{ESL_{l',l}}$, which captures
the opposite confusion. A similarity estimate for a language pair is then obtained
by averaging these two scores: 
\begin{equation}
\small
S_{ESL_{l,l'}}=S_{ESL_{l',l}}=\frac{1}{2}(S'_{ESL_{l,l'}}+S'_{ESL_{l',l}})
\end{equation}
As shown in \cite{berzak2014}, given the similarity matrix $S_{ESL}$, one can obtain an approximation for the 
typology of a native language by projecting the typological features from its most
similar languages. Here, we use the typology of the closest language, an approach 
that yields 70.7\% accuracy in predicting the typological features of our set of languages.

In the bootstrapping setup, we train the regression models on the true typology
of the languages in the training set, and use the approximate typology of the test language to
predict the relative error rates of its speakers in ESL. 

\subsection{Results}

Table \ref{aprox_results} summarizes the error prediction results using 
approximate typological features for the test languages. As can be seen,
our approach continues to provide substantial performance 
gains despite the inaccuracy of the typological information used for the test
languages. 
The best performing method, \emph{RegCA} reduces
the \emph{MAE} of \emph{Base} by 13.9\%, with performance improvements for most of the
languages and error types. Performance gains are also obtained on the $D_{KL}$
metric, whereby \emph{RegCA} scores 0.041, compared to the \emph{Base} score of 0.052, 
improving on 11 out of our 14 languages.

{\renewcommand{\arraystretch}{1.1}
\begin{table}[h!]
\footnotesize
\begin{center}
\begin{tabular}{|l|l|l|l|l|}
\hline
\bf                   & \bf Base & \bf NN & \bf Reg & \bf RegCA  \\ \hline
 \textbf{MAE}  & 1.28 & 1.12 & 1.13 & \textbf{1.10}  \\ \hline 
 \textbf{Error Reduction}   &   -  & 12.6   & 11.6        &\textbf{13.9} \\ \hline \hline
 \#Languages   &   -        & \textbf{11/14} &  \textbf{11/14} &  \textbf{11/14} \\ \hline 
 \#Mistakes    &   -  & 10/20 & 10/20& \textbf{11/20} \\ \hline \hline
 \textbf{AVG $\mathbf{D_{KL}}$}  & 0.052 & 0.048 & 0.043 & \textbf{0.041} \\ \hline 
 \#Languages   &   -  & 10/14 & \textbf{11/14} & \textbf{11/14} \\ \hline 

\end{tabular}
\end{center}
\caption{Results for prediction of relative error frequencies using the bootstrapping approach. In this setup, the true typology of the test language
is substituted with approximate typology derived from morpho-syntactic ESL features. 
	}\label{aprox_results}
\end{table}
}

\section{Conclusion and Future Work}\label{sec:conclusion}

We present a computational framework for predicting native language specific grammatical 
error distributions in ESL, based on the typological properties 
of the native language and their compatibility with the typology of 
English. Our regression model achieves substantial performance improvements as compared to a 
language oblivious baseline, as well as a language dependent nearest neighbor baseline. Furthermore, 
we address scenarios in which the typology of the native language 
is not available, by bootstrapping typological features from ESL texts.
Finally, inspection of the model parameters 
allows us to identify native language properties which play
a pivotal role in generating different types of grammatical errors.

In addition to the theoretical contribution, the outcome of our work has a strong potential
to be beneficial in practical setups.
In particular, it can be utilized for developing educational
curricula that focus on the areas of difficulty that are characteristic
of different native languages. Furthermore, the derived error frequencies 
can be integrated as native language specific priors in 
systems for automatic error correction. 
In both application areas, previous work relied on the existence of 
error tagged ESL data for the languages of interest. Our approach
paves the way for addressing these challenges even in the absence
of such data.

\section*{Acknowledgments}
This material is based upon work supported by the Center for Brains, Minds, 
and Machines (CBMM), funded by NSF STC award CCF-1231216.

\bibliographystyle{acl}
\bibliography{acl2015ca}

\begin{thebibliography}{}

\bibitem[\protect\citename{Berzak \bgroup et al.\egroup }2014]{berzak2014}
Yevgeni Berzak, Roi Reichart, and Boris Katz.
\newblock 2014.
\newblock Reconstructing native language typology from foreign language usage.
\newblock In {\em Proceedings of the Eighteenth Conference on Computational
  Natural Language Learning}, pages 21--29. Association for Computational
  Linguistics, June.

\bibitem[\protect\citename{Daum{\'e}~III}2009]{daume2009clustering}
Hal Daum{\'e}~III.
\newblock 2009.
\newblock Non-parametric {B}ayesian areal linguistics.
\newblock In {\em Proceedings of human language technologies: The 2009 annual
  conference of the north american chapter of the association for computational
  linguistics}, pages 593--601. Association for Computational Linguistics.

\bibitem[\protect\citename{de Marneffe \bgroup et al.\egroup
  }2006]{2006stanforddep}
Marie-Catherine de~Marneffe, Bill MacCartney, Christopher~D Manning, et~al.
\newblock 2006.
\newblock Generating typed dependency parses from phrase structure parses.
\newblock In {\em Proceedings of LREC}, volume~6, pages 449--454.

\bibitem[\protect\citename{Fries}1945]{fries1945}
Charles~C Fries.
\newblock 1945.
\newblock Teaching and learning english as a foreign language.

\bibitem[\protect\citename{Gass and Selinker}1992]{gass1992language}
Susan~M Gass and Larry Selinker.
\newblock 1992.
\newblock {\em Language Transfer in Language Learning: Revised edition},
  volume~5.
\newblock John Benjamins Publishing.

\bibitem[\protect\citename{Georgi \bgroup et al.\egroup
  }2010]{georgi2010clustering}
Ryan Georgi, Fei Xia, and William Lewis.
\newblock 2010.
\newblock Comparing language similarity across genetic and typologically-based
  groupings.
\newblock In {\em Proceedings of the 23rd International Conference on
  Computational Linguistics}, pages 385--393. Association for Computational
  Linguistics.

\bibitem[\protect\citename{Jarvis and Pavlenko}2007]{jarvis2007transfer}
Scott Jarvis and Aneta Pavlenko.
\newblock 2007.
\newblock {\em Crosslinguistic influence in language and cognition}.
\newblock Routledge.

\bibitem[\protect\citename{Kruskal and Wallis}1952]{kruskal1952}
William~H Kruskal and W~Allen Wallis.
\newblock 1952.
\newblock Use of ranks in one-criterion variance analysis.
\newblock {\em Journal of the American statistical Association},
  47(260):583--621.

\bibitem[\protect\citename{Lado}1957]{lado1957}
Robert Lado.
\newblock 1957.
\newblock Linguistics across cultures: Applied linguistics for language
  teachers.

\bibitem[\protect\citename{Levene}1960]{levene1960}
Howard Levene.
\newblock 1960.
\newblock Robust tests for equality of variances.
\newblock {\em Contributions to probability and statistics: Essays in honor of
  Harold Hotelling}, 2:278--292.

\bibitem[\protect\citename{Mann and Whitney}1947]{mann1947test}
Henry~B Mann and Donald~R Whitney.
\newblock 1947.
\newblock On a test of whether one of two random variables is stochastically
  larger than the other.
\newblock {\em The annals of mathematical statistics}, pages 50--60.

\bibitem[\protect\citename{Nagata and Whittaker}2013]{nagata2013}
Ryo Nagata and Edward Whittaker.
\newblock 2013.
\newblock Reconstructing an indo-european family tree from non-native english
  texts.
\newblock In {\em Proceedings of the 51st Annual Meeting of the Association for
  Computational Linguistics}, pages 1137--1147, Sofia, Bulgaria. Association
  for Computational Linguistics.

\bibitem[\protect\citename{Nicholls}2003]{nicholls2003clc}
Diane Nicholls.
\newblock 2003.
\newblock The cambridge learner corpus: Error coding and analysis for
  lexicography and elt.
\newblock In {\em Proceedings of the Corpus Linguistics 2003 conference}, pages
  572--581.

\bibitem[\protect\citename{Odlin}1989]{odlin1989transfer}
Terence Odlin.
\newblock 1989.
\newblock {\em Language transfer: Cross-linguistic influence in language
  learning}.
\newblock Cambridge University Press.

\bibitem[\protect\citename{Rozovskaya and Roth}2011]{rozovskaya2011}
Alla Rozovskaya and Dan Roth.
\newblock 2011.
\newblock Algorithm selection and model adaptation for esl correction tasks.
\newblock In {\em Proceedings of the 49th Annual Meeting of the Association for
  Computational Linguistics: Human Language Technologies-Volume 1}, pages
  924--933. Association for Computational Linguistics.

\bibitem[\protect\citename{Rozovskaya and Roth}2014]{rozovskaya2014}
Alla Rozovskaya and Dan Roth.
\newblock 2014.
\newblock Building a state-of-the-art grammatical error correction system.
\newblock {\em Transactions of the Association for Computational Linguistics},
  2(10):419--434.

\bibitem[\protect\citename{Shapiro and Wilk}1965]{shapiro1965}
Samuel~Sanford Shapiro and Martin~B Wilk.
\newblock 1965.
\newblock An analysis of variance test for normality (complete samples).
\newblock {\em Biometrika}, pages 591--611.

\bibitem[\protect\citename{Swanson and Charniak}2013]{swanson2013}
Ben Swanson and Eugene Charniak.
\newblock 2013.
\newblock Extracting the native language signal for second language
  acquisition.
\newblock In {\em HLT-NAACL}, pages 85--94.

\bibitem[\protect\citename{Swanson and Charniak}2014]{swanson2014}
Ben Swanson and Eugene Charniak.
\newblock 2014.
\newblock Data driven language transfer hypotheses.
\newblock {\em EACL 2014}, page 169.

\bibitem[\protect\citename{Tetreault \bgroup et al.\egroup
  }2013]{nli2013report}
Joel Tetreault, Daniel Blanchard, and Aoife Cahill.
\newblock 2013.
\newblock A report on the first native language identification shared task.
\newblock In {\em Proceedings of the Eighth Workshop on Innovative Use of NLP
  for Building Educational Applications}, pages 48--57. Citeseer.

\bibitem[\protect\citename{Wardhaugh}1970]{wardhaugh1970}
Ronald Wardhaugh.
\newblock 1970.
\newblock The contrastive analysis hypothesis.
\newblock {\em TESOL quarterly}, pages 123--130.

\bibitem[\protect\citename{Whitman and
  Jackson}1972]{whitman1972unpredictability}
Randal~L Whitman and Kenneth~L Jackson.
\newblock 1972.
\newblock The unpredictability of contrastive analysis.
\newblock {\em Language learning}, 22(1):29--41.

\bibitem[\protect\citename{Wong and Dras}2009]{wong2009contrastive}
Sze-Meng~Jojo Wong and Mark Dras.
\newblock 2009.
\newblock Contrastive analysis and native language identification.
\newblock In {\em Proceedings of the Australasian Language Technology
  Association Workshop}, pages 53--61. Citeseer.

\bibitem[\protect\citename{Yannakoudakis \bgroup et al.\egroup
  }2011]{fcecorpus2011}
Helen Yannakoudakis, Ted Briscoe, and Ben Medlock.
\newblock 2011.
\newblock A new dataset and method for automatically grading {ESOL} texts.
\newblock In {\em ACL}, pages 180--189.

\end{thebibliography}

\end{document}